\documentclass[10pt, a4paper, onecolumn]{article}

\usepackage[numbers]{natbib}

\usepackage[T1]{fontenc}
\usepackage{lmodern}

\usepackage{natbib}
\usepackage[utf8]{inputenc} 
\usepackage{booktabs}       
\usepackage{amsfonts}       
\usepackage{nicefrac}       
\usepackage{microtype}      
\usepackage{xcolor}         
\usepackage{graphicx}
\usepackage{amsmath,amssymb,amsfonts}
\usepackage{bm}
\usepackage{url}

\usepackage{array,multirow,graphicx}

\usepackage{geometry}
\newgeometry{vmargin={15mm}, hmargin={12mm,17mm}}

\usepackage{multirow}
\usepackage{multicol}

\usepackage{algorithm}
\usepackage[noend]{algpseudocode}

\title{A Trainable Feature Extractor Module for Deep Neural Networks and Scanpath Classification}

\author{
	Wolfgang Fuhl
	University Tübingen\\
	Tübingen, 72076 \\
	\texttt{wolfgang.fuhl@uni-tuebingen.de} \\
}

\begin{document}
	
	\maketitle
	
	\begin{abstract}
		Scanpath classification is an area in eye tracking research with possible applications in medicine, manufacturing as well as training systems for students in various domains. In this paper we propose a trainable feature extraction module for deep neural networks. The purpose of this module is to transform a scanpath into a feature vector which is directly useable for the deep neural network architecture. Based on the backpropagated error of the deep neural network, the feature extraction module adapts its parameters to improve the classification performance. Therefore, our feature extraction module is jointly trainable with the deep neural network. The motivation to this feature extraction module is based on classical histogram-based approaches which usually compute distributions over a scanpath. We evaluated our module on three public datasets and compared it to the state of the art approaches.
	\end{abstract}

	\section{Introduction}
	
	\begin{figure}
		\centering
		\includegraphics[width=\textwidth]{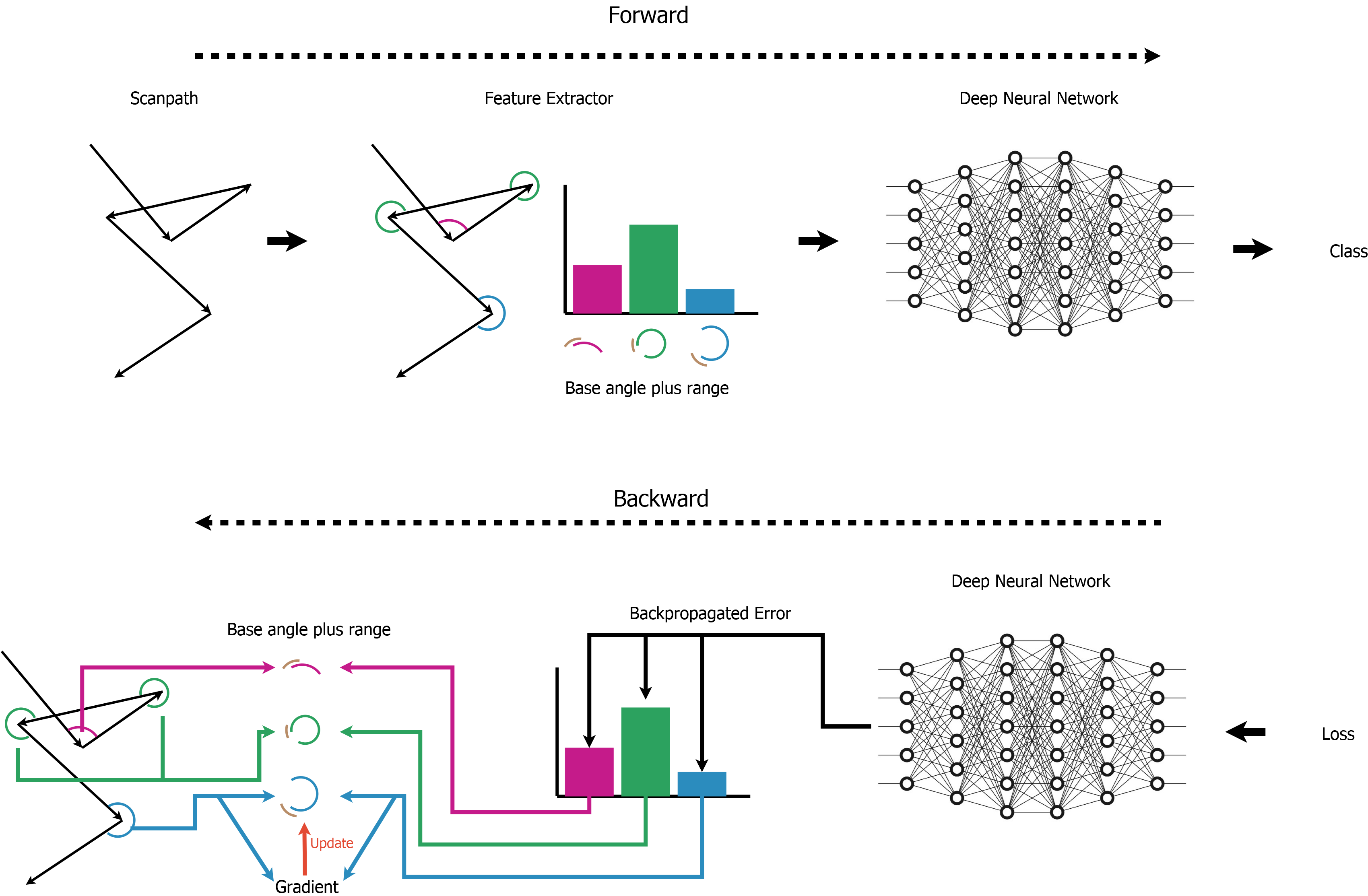}
		\caption{The forward and backward pass of the proposed feature extractor module.}
		\label{fig:teaser}
	\end{figure}

	Our eye movements reveal more than just what we see. They also show how our brain and senses work together. By looking at the sequence and duration of eye fixations and jumps, called the scanpath, we can learn how we process information. In different fields, eye tracking studies have discovered patterns in eye movements. These patterns can distinguish between different types of people (e.g., beginners and experts), or different situations, such as the task given to a person. For example in art, eye movement differences have been observed between professional and novice art viewers for both realistic and abstract art~\cite{zangemeister1995evidence}. Moreover, top-down expectations and bottom-up visual features can influence the eye movements on artworks ~\cite{locher2015art,massaro2012art}. Similarly in the medical field, scanpath differences can indicate the professional and the treatment factors. Scanpath differences between beginners and experts have been reported in microneurosurgeons ~\cite{eivazi2012gaze,kubler2017subsmatch} and radiologists ~\cite{manning2006radiologists,van2017radiologists}. It was also found that dental students who took a specific radiography training course could be correctly identified from their scanpaths.
	
	Regarding the treatment factor, eye movement differences from healthy controls have been observed in both patients with schizophrenia ~\cite{hooker2005you} and autism spectrum disorder ~\cite{nation2008sensitivity,volkmar1990gaze}. Therefore, scanpaths can potentially be used for more precise training, diagnosis, and treatment methods. In driving, scanpaths have been applied to reliably assess safe or unsafe driving in people with visual impairments ~\cite{kubler2017subsmatch}. Moreover, they can be utilized in driver assistance systems to signal the take-over readiness ~\cite{vicente2015driver} or cognitive load ~\cite{krejtz2018eye}, and fatigue~\cite{wang2019eye}. Notably, most of the studies above focus on finding statistically significant differences in single scanpath measures. Hence, there is a large and ever increasing body of scanpath comparison and classification methodology: From simple statistics to state-of-the-art machine learning~\cite{byrne2023exploring}.
	
	In this paper we propose a novel deep neural network layer for scanpath classification which is inspired by the approach from ~\cite{fuhl2019ferns}. Our approach is not dependent on generated or predefined areas of interest and works on the gaze samples directly. We reformulated the angle and angle range approach from ~\cite{fuhl2019ferns} in a way that it is trainable by the backpropagation algorithm. This means that our novel layer for deep neural networks is jointly trainable and works as a feature extraction module in front of the classification part, which is usually a fully connected stage.
	
	In short, our contributions are:
	\begin{itemize}
		\item A novel feature extraction layer for deep neural networks.
		\item Integration of the angle and angle range approach into the backpropagation algorithm.
		\item Evaluations on multiple public datasets and comparison to the state of the art approaches.
	\end{itemize}
		
	\section{Related work}
	In the 1990s, the initial proposal of automated metrics took place~\cite{brandt1997spontaneous}. Since then, there has been a significant evolution in the methodology for automated scanpath comparison~\cite{anderson2015comparison}. Recently, machine-learning based approaches have emerged, showing impressive results~\cite{crabb2014s,french2017evaluation,hoppe2018eye,zhang2018old}. These approaches are capable of distinguishing relevant eye movement patterns from high levels of noise. However, there is still an ongoing debate regarding how to efficiently encode eye movement trajectories for machine learning purposes. Some algorithms heavily rely on time-aggregated features or complete-sequence alignment, while others focus on gaze transitions (i.e., the shift of gaze between two targets, also known as a saccade) as a popular feature~\cite{hoppe2018eye,cristino2010scanmatch,dewhurst2018task,burch2018eyemsa}. This strategy allows for the modeling of cognitive associations between gaze targets. Hidden Markov Models (HMMs) remained long the most common approach~\cite{coutrot2018scanpath,hacisalihzade1992visual}, but there are also other methods that extend these patterns to span multiple subsequent fixations and saccades. Identifying patterns in longer sequences is particularly important, as these patterns can be more specific to a task or subject group, making them highly useful for classification~\cite{kubler2017subsmatch}.
	
	A CNN related to autism spectrum disorder used fixation maps from the entire scanpath as input~\cite{elbattah2019learning}. Multiple deep learning models for schizophrenia and task classification also used heatmaps as input~\cite{kacur2020analysis,vortmann2021imaging}. To include temporal information, CNN-LSTM networks classified autism spectrum disorder using scanpath-based patches from a saliency map~\cite{tao2019sp}. \cite{elbattah2019learning} used gaze snapshots based on group attention at time intervals as input for an autoencoder. These attention map methods often apply techniques like Gaussian blurring to the raw data. Other methods use different scanpath image creations for classifications, such as Markov random fields~\cite{wang2015encoding}, object detection based~\cite{venuprasad2020analyzing}, input image patch based~\cite{castner2020deep}, and principle component based ~\cite{kumar2020challenges}.
	
	Creating images from the raw scanpath data is another representation option. This approach preserves potentially relevant information for the model that pre-processing could remove. \cite{sims2020neural} used scanpath images from the raw gaze for the entire duration and for five second intervals as input for an RNN to classify confusion. \cite{ahmed2020convolutional} used scanpath images by connecting saccades and weighting them by the fixation densities with a CNN. A generative model for scanpath classification that converted gaze data into emojis was proposed in \cite{fuhl2019encodji}. It encodes gaze data as a compact image with the red, green, and blue channels representing the spatial, temporal, and connectivity which is than fed to a deep neural network. \cite{atyabi2023stratification,bhattacharya2020relevance,byrne2023exploring} explored different scanpath representations for classification. They used temporal coloring for saccade velocities or symbols for different fixation durations.

	\section{Method}
	Figure~\ref{fig:teaser} shows the workflow of the proposed approach, which is inspired by the random ferns used in \cite{fuhl2019ferns}. In the forward pass, our layer gets the entire sequence and checks a series of inter-sample angles and angle ranges. Based on these checks, a histogram index is selected and increased. The final histograms are given to a classifier, which is, in our case, a residual neural network with convolutions. This part of our approach is similar to what was done in the original paper with balanced decision trees, also known as random ferns \cite{fuhl2019ferns} without the residual network as a classifier. The interesting part and the main contribution of our method is the backward pass. Here, we propagate the gradient back to each histogram bin. Based on the sign of the gradient, we either increase or decrease the angle range. In the following, we will describe our module and its integration into the backpropagation algorithm in detail.
	
	\begin{algorithm}
		\caption{The algorithmic description of the forward pass for the proposed module. First, we set the histograms to zero. Afterward, all angle and angle range checks are applied to the entire sequence. Based on these checks, an index for the current histogram that belongs to a set of angles and angle ranges is computed. With this index, a histogram bin is increased, and in the end, each histogram is normalized.}
		\label{alg:forward}
		\begin{algorithmic}[1]
			\Procedure{ForwardPass}{$sequence, anglesets, histograms$}       
			\State $histograms=0$
			\ForAll{$set_i \in anglesets$}
			\For{$seq_j=0; seq_j<size(sequence-size(set_i)); seq_j++$}
			\State $index=0$
			\ForAll{$angleANDrange_k \in set_i$}
			\If{$angleANDrange_k(sequence(seq_j+k))$}	\Comment{1 if in range}
			\State $index+=2^k$
			\EndIf
			\EndFor
			\State $histograms[index(set_i)][index]+=1$
			\EndFor
			\For{$histo_i=0; seq_j<size(set_i); seq_j++$}	\Comment{Normalize histogram}
			\State $histograms[index(set_i)][histo_i]=\frac{histograms[index(set_i)][histo_i]}{\sum_{l=1}^{size(set_i)} histograms[index(set_i)][l]}$
			\EndFor
			\EndFor
			\EndProcedure
		\end{algorithmic}
	\end{algorithm}
	
	Algorithm~\ref{alg:forward} describes the forward pass of our approach in the backpropagation algorithm. Before we can use our layer in a neural network we have to specify two parameters, one is the amount of angle sets which is the amount of sequences consisting of angles and angle ranges we want to use. The second parameter is the length of such a sequence. A simplified illustration can be found in Figure~\ref{fig:teaser}. In this illustration, the sequence length would be one, which means that each angle and angle range has its own bin in the histogram. After we set up those two parameters, we evaluate each angle and angle range sequence on the given samples from an eye tracking recording and compute the histograms as described in Algorithm~\ref{alg:forward}. The filled histograms are normalized and given to a neural network for further processing.
	
	\begin{algorithm}
		\caption{The backward pass of our approach is described as an algorithm. In the first part, we have to do the forward pass again. With the computed indexes, we can access the gradients corresponding to different angle and angle range sets. Next, we need to compute which angle and angle range check are evaluated positive. For all positive angle and angle range checks, we sum up the corresponding gradients without updating them directly. In the last step, we update all angle ranges according to the cumulated gradients. We do not update them directly since this would change the angle and angle range checks, which would invalidate our gradient computation. In the real implementation, we do not need to compute the indexes since they are stored with the forward pass. In addition, we also do not need to evaluate which angle and angle range check evaluated positive, since this is already known based on the histogram index. This means that we described the backward pass in a way it can be computed and that can be understood more easily. The real implementation therefore differs from the algorithm to save resources and training time.}
		\label{alg:backward}
		\begin{algorithmic}[1]
			\Procedure{BackwardPass}{$sequence, anglesets, gradient, learningrate$}       
			\ForAll{$set_i \in anglesets$}
			\For{$seq_j=0; seq_j<size(sequence-size(set_i)); seq_j++$}
			\State $index=0$
			\ForAll{$angleANDrange_k \in set_i$}
			\If{$angleANDrange_k(sequence(seq_j+k))$}	\Comment{1 if in range}
			\State $index+=2^k$
			\EndIf
			\EndFor
			\ForAll{$angleANDrange_k \in set_i$}
			\If{$angleANDrange_k(sequence(seq_j+k))$}	\Comment{1 if in range}
			\State $angleANDrange_k.rangeUpdate+=gradient[i][index]$
			\State	\Comment{Computation of the cumulative update of the angle range}
			\EndIf
			\EndFor
			\EndFor
			\EndFor
			\ForAll{$set_i \in anglesets$}
			\ForAll{$set_i \in anglesets$}
			\ForAll{$angleANDrange_k \in set_i$}
			\State $angleANDrange_k.range+=angleANDrange_k.rangeUpdate*learningrate$
			\State	\Comment{Applying the cumulative update to the angle range}
			\EndFor
			\EndFor
			\EndFor
			\EndProcedure
		\end{algorithmic}
	\end{algorithm}
	
	Algorithm~\ref{alg:backward} describes the backward pass of our approach. For simplification, we added the parts of the forward pass into it so it can be fully understood. In the real implementation, we store the result of the forward pass, and we also know based on the histogram bin which angle and angle range evaluated to one due to the way we set them up in our memory. The input to the neural network, our histograms, receive the back propagated error from the network. With the error or gradient assigned to each bin in our histograms, we can compute which angle and which angle range has evaluated to one. The corresponding angle ranges are then adjusted based on the gradient. If the gradient is negative, we reduce the angle range and if it is positive, we increase the angle range. Since multiple angle ranges participated on multiple places over a sequence, we accumulate the gradient first. This is indicated by ".rangeUpdate" in Algorithm~\ref{alg:backward}.
	
	\section{Evaluation}
	In this section we first describe the used public datasets and how we performed the training, validation and test splits. Afterward, the training parameters of our approach and the configuration of the other approaches is described. The last part in this section presents and discusses our results.

	\subsection{Datasets}
	Gaze~\cite{dorr2010variability}: A data set with eye tracking data on moving scenes. The data was collected using an SR Research EyeLink II eye tracker with 250 Hz. For our experiment, we used the data given for static images where each static image of a video was treated as the same image. Moreover, we omitted subject V01 since there was only one recording available. Hence, we used the eye tracking data of 10 subjects on 9 images for our experiment with an average recording duration of 2 seconds. The training and test split was done using 50\% for the training and 50\% for the testing with a random selection. For the stimulus classification we made sure that no subject is shared between the training and testing set and for the subject classification we did the same based on the stimulus.

	WherePeopleLook~\cite{judd2009learning}: An eye tracking dataset that focuses on integrating top-down features into the generation of saliency maps. This dataset comprises 1003 static images, each accompanied by eye tracking data from 15 subjects. The eye tracking data was collected for an average recording length of 3 seconds per image. To conduct our experiment, we divided the dataset into a training set and a testing set, ensuring a balanced split of 50\% for each. We took great care to ensure that no subject appeared in both the training and testing sets for stimulus classification. Similarly, for subject classification, we made sure that no stimulus overlapped between the training and testing sets. This meticulous approach guarantees the integrity and reliability of our experiment results.

	DOVES~\cite{rajashekar2009doves}: An extensive eye tracking dataset consisting of data from 29 subjects recorded on 101 natural images. The recordings were conducted using a high-precision dual-Purkinje eye tracker with a sampling rate of 200 Hz. Each recording had an average length of 5 seconds. Following the approach used in the WherePeopleLook dataset, we split the dataset into training and testing sets, with an equal distribution of 50\% for each. To ensure accurate stimulus classification, we took care to avoid any overlap of subjects between the training and testing sets. Similarly, for subject classification, we ensured that no stimulus was shared between the training and testing sets. This rigorous methodology ensures the reliability and validity of our dataset for further analysis and experimentation.
	
	We decided to use those datasets since they have a large amount of available sequences, which is important for neural network based approaches. 
	
	\subsection{Training and adaption of the other methods}
	\label{sec:trainadapt}
	All used CNNs (Convolution neural networks) are ResNet-12~\cite{he2016deep} in our evaluation. This concerns all indications of "+ CNN" in Table~\ref{tbl:evalall} and our approach. For the training of those networks, we used a soft max classification layer with the stochastic gradient descent optimizer and momentum. The initial learning rate was set to $10^{-3}$ and reduced to $10^{-4}$ after 50 epochs. With the learning rate of $10^{-4}$ we trained additional 50 epochs and used the best model based on the results of the validation set which consists of 20\% randomly selected from our training set. For the features HOV and HEAT we computed multiple parameter configurations and selected the best model based on the 20\% validation set. For \cite{byrne2023exploring} we evaluated all representations and selected the best performing representation based on the results on the 20\% validation set. For *RNN~\cite{sims2020neural} and *LSTM~\cite{tao2019sp} we did not find the code online and therefore tried to reproduce the approach as best as we could. For Subsmatch 2.0 we selected the best performing AOI selection approach, and we also tried different classifiers for the features as well as different parameters for the classifiers. For the random fern approach, we evaluated different feature selection and ensemble combination parameters and selected the best performing one. For the *EM Statistics and Auto AOI + statistics we used duration, speed, and acceleration based on the AOIs or the eye movement types which were classified using a velocities and dispersion threshold. Based on duration, speed, and acceleration, we computed the mean, variance, standard deviation, and the confidence intervals. All together was one feature vector. The deep semantic gaze embedding was also reproduced with a ResNet-12. For the Encodji approach, we used the ResNet-12 as classifier as well as discriminator to train the generative adversarial network. The generative adversarial network itself was a U-Net with interconnections. All approaches had the same training, validation and testing data.

	\subsection{Results}
	
	\begin{table*}
		\centering
		\caption{The metric in this table is accuracy rounded to two decimal places. We evaluated different initializations of our approach for the dataset Gaze. We performed the subject and stimulus classification. Angle sets are the amount of all sequences of angles and angle ranges that are randomly initialized and evaluated during the forward pass. Set size is the amount of angles and angle ranges that are in one angle set. For the evaluation we performed a 80\% training and 20\% validation split on the training data only.}
		\label{tbl:paramEval}
		\begin{tabular}{cccc}
			& & \multicolumn{2}{c}{Dataset Gaze}\\ 
			Angle sets & Set size & 10 Classes Subject & 9 Classes Stimulus \\ \hline
			$2^{9}$ & 4 & 54 &  22 \\ 
			$2^{9}$ & 5 & 68 & 31 \\ 
			$2^{9}$ & 6 & 71 & 34 \\ 
			$2^{10}$ & 4 & 73 & 39  \\ 
			$2^{10}$ & 5 & 82 &  51 \\ 
			$2^{10}$ & 6 & 81 &  47 \\ 
			$2^{11}$ & 4 & 78 &  43 \\ 
			$2^{11}$ & 5 & 89 &  58 \\ 
			$2^{11}$ & 6 & 87 &  55 \\ 
			$2^{12}$ & 4 & 82 &  49 \\ 
			$2^{12}$ & 5 & \textbf{91} & \textbf{62} \\ 
			$2^{12}$ & 6 & 88 &  57 \\ 
		\end{tabular}
	\end{table*}
	
	In Table~\ref{tbl:paramEval} we evaluated different combinations of our two parameters amount of angle sets and the set size. For the evaluation, we used 80\% of the training data and 20\% of the training data for validation. The reported accuracy is rounded and computed on the validation set. As can be seen, the amount of sequences has a huge impact on the classification accuracy. This is the case since randomly selected angle and angle range sequences can also be useless. With more such sequences, we increase the chance in getting good combinations. For the set size parameter the same is true since larger sequences have a lower probability of a good selection since there are more possible combinations. This means that both parameters are in relation to each other, which means that higher set sizes also require larger amounts of angle sets. Based on our evaluation, we have selected $2^{12}=4.096$ angle sets and an angle as well as angle range sequence length of 4. For binary decisions, end up in $2^{5}=32$ possible combinatorial outcomes. Therefore, our produced tensor for the neural network has a size of $4.096 \times 32$.

	\begin{table*}
		\centering
		\caption{The used metric is accuracy, which we rounded to two decimal places. We compared our approach with the best parameters from Table~\ref{tbl:paramEval} with other state of the art approaches. Best results are shown in bold, and * indicates that we have reimplemented the methods as it is described in Section~\ref{sec:trainadapt}.}
		\label{tbl:evalall}
		\begin{tabular}{lcccccc}
			Dataset & \multicolumn{2}{c}{Gaze} & \multicolumn{2}{c}{WherePeopleLook}& \multicolumn{2}{c}{DOVES}\\ 
			Target & Sub & Stim & Sub & Stim & Sub & Stim \\ 
			Classes & 10 & 9  & 15 & 1003  & 29 &  101 \\ \hline
			*Deep semantic gaze embedding~\cite{castner2020deep} & 80 & 38 & 39 & 37 & 15 & 48 \\  
			Random Ferns~\cite{fuhl2019ferns} & 85 & 44 & \textbf{42} & \textbf{41} & 18 & \textbf{52} \\  
			Subsmatch 2.0~\cite{kubler2017subsmatch} & 69 & 28 & 27 & 30 & 8 & 30 \\  
			*EM Statistics~\cite{goldberg2010visual} + SVM & 61 & 22 & 21 & 20 & 5 & 28 \\ 
			*EM Statistics~\cite{goldberg2010visual} + Tree ENS & 62 & 24 & 25 & 21 & 6 & 31 \\  
			*EM Statistics~\cite{goldberg2010visual} + two layer NN & 60 & 24 & 23 & 21 & 6 & 29 \\  
			Auto AOI~\cite{fuhl2023area} + statistics + SVM & 69 & 27 & 29 & 27 & 9 & 39 \\  
			Auto AOI~\cite{fuhl2023area} + statistics + Tree ENS & 72 & 29 & 30 & 28 & 10 & 41 \\  
			Auto AOI~\cite{fuhl2023area} + statistics + two layer NN & 71 & 29 & 29 & 28 & 10 & 40\\  
			Encodji~\cite{fuhl2019encodji} & 83 & 39 & 36 & 34 & 16 & 48 \\   
			Encodji input only~\cite{fuhl2019encodji} + CNN & 77 & 31 & 32 & 31 & 11 & 45 \\  
			Best from ~\cite{byrne2023exploring} + CNN & 78 & 33 & 31 & 33  & 12 & 47 \\  
			HEAT + SVM~\cite{fuhl2021gaze} & 74 & 28 & 30 & 29 & 8 & 46 \\  
			HEAT + Tree ENS~\cite{fuhl2021gaze} & 76 & 29 & 33 & 31 & 10 & 47 \\  
			HEAT + two layer NN~\cite{fuhl2021gaze} & 75 & 26 & 30 & 28 & 9 & 44 \\  
			HOV + SVM~\cite{fuhl2018histogram} & 74 & 31 & 31 & 28 & 10 & 45  \\  
			HOV + Tree ENS~\cite{fuhl2018histogram} & 75 & 30 & 33 & 30 & 11 & 46 \\  
			HOV + two layer NN~\cite{fuhl2018histogram} & 73 & 27 & 31 & 26 & 9 & 43 \\  	
			*RNN~\cite{sims2020neural} & 70 & 26 & 27 & 22 & 6 & 31 \\
			*LSTM~\cite{tao2019sp} & 67 & 23 & 26 & 20 & 5 & 27 \\  \hline
			Proposed & \textbf{87} & \textbf{45} & \textbf{42} & 40 & \textbf{19} & \textbf{52} \\ 
		\end{tabular}
	\end{table*}
	
	Table~\ref{tbl:evalall} shows the comparison to other state of the art approaches with the accuracy metric. As can be seen, our approach works similarly well as the random fern approach from which we used the feature extraction approach. This means that the sequence of angles and angle ranges seams to be a good feature for the used datasets or scanpath in general. What can be seen also is that all neural network based approaches, excluding the ones on statistics, worked well on the datasets. The reason for this is possibly that the large amount of available sequences helped the neural networks. The statistical based approaches as well as the features HEAT and HOV worked not so well on the datasets, which is due to the short recording length. Statistics and features like the histogram of oriented gradients or the heatmaps need possibly more data to be discriminative. The worst approaches were the RNN and the LSTM, they suffer the most from the short recording length since they require long sequences for which they are designed. In total, our approach was at least as good as the random ferns. Sometimes our approach outperformed the ferns, and for the stimulus classification in the WherePeopleLook dataset the ferns were slightly better compared to our approach. Therefore, we think our approach to integrate the angle and angle range feature into deep neural networks or into the backpropagation algorithm was successful.
	
	\section{Limitations}
	We compared our approach on three public datasets with different amounts of subjects and stimuli. While this can be seen as an extensive evaluation, it is not guaranteed that the results apply for all datasets. One example here are long term recordings in real life which are publicly not available as a dataset, which is also the reason why we did not evaluate on such datasets. Another limitation of our paper are the optimal parameters for the state of the art approaches. We tried to reproduce the methods and approaches as good as we could and also tried to select the best parameters, but it is still possible that there are preprocessing steps or parameter combinations which lead to better result.
	
	\section{Conclusion and Outlook}
	In this paper, we proposed a module for deep neural networks which is based on the angle and angle range approach from \cite{fuhl2019ferns}. Our main contribution is the integration of the approach into the backpropagation algorithm, which makes it possible to train it jointly with deep neural networks or any other computational graph based approach which requires derivatives for gradient determination. Our approach outperformed most of the state of the art approaches, but it has to be noted that all of our datasets have only a small recording length since long term recordings with many subjects and many sequences of scanpath are not publicly available, which is especially true for medical recordings. The approach with random ferns is very close to our results, and sometimes beats our deep neural network based approach. This means that it is still the case that entropy and information gain, which is used to train decision trees like the random ferns, can outperform deep neural networks as it is common for tabular data. In addition, it is also obvious that especially for the used datasets, the consecutive sample angles form strong features for the classification. Overall we think that our proposed method to integrate the consecutive angle and angle range approach into the backpropagation algorithm was successful since it delivers results close or slightly better than the original approach based on decision trees~\cite{fuhl2019ferns}. Future work should investigate if it is possible to also learn the base angle itself as well as more advanced module for an internal use in deep neural networks. In addition, it could be possible to use it in generative adversarial networks to generate human visual behavior.

	\section{Potentially harmful impacts and future societal risks}
	The proposed approach is directed into the research area of eye tracking with the purpose to help humans in terms of a supportive diagnosis system or to be part of educational software. Of course, there are many possibilities to use scanpath classification in harmful ways, like the observation and intention prediction of humans. The classification of abnormal behavior could for example reveal if somebody has a sickness like Alzheimer or Autism, which is private information. Scanpath analysis could also be used to train humans for specific harmful tasks or to cheat in competitions like poker for example. We as researchers do not want our knowledge to be used in such areas or for such tasks, but we cannot prevent it from happening.

	\section*{Acknowledgement}
	Funded by the Deutsche Forschungsgemeinschaft (DFG, German Research Foundation) – 508330921
	
	\bibliographystyle{plain}
	\bibliography{template}

\end{document}